\documentclass{article}

\usepackage[main, final, nonatbib]{neurips_2026}

\usepackage[utf8]{inputenc}
\usepackage[T1]{fontenc}
\usepackage{hyperref}
\usepackage{hyperxmp}
\hypersetup{linkbordercolor={0.494 0.737 0.698},citebordercolor={0.494 0.737 0.698},urlbordercolor={0.494 0.737 0.698}}
\usepackage{booktabs}
\usepackage{amsfonts}
\usepackage{nicefrac}
\usepackage{microtype}
\usepackage{xcolor}

\usepackage{float}
\usepackage{wrapfig}
\usepackage{algorithm}
\usepackage{graphicx}
\usepackage{subcaption}
\usepackage{amsmath}
\usepackage{amssymb}
\usepackage{multirow}
\usepackage{makecell}
\usepackage{enumitem}
\usepackage{fancyvrb}

\usepackage[style=numeric-comp,sorting=none,maxbibnames=3,minbibnames=1,url=false,doi=false,isbn=false,date=year]{biblatex}
\DeclareSourcemap{\maps[datatype=bibtex]{\map{\step[fieldset=file, null]\step[fieldset=abstract, null]\step[fieldset=annotation, null]\step[fieldset=annote, null]\step[fieldset=keywords, null]}}}
\DeclareFieldFormat[article,misc,inproceedings,book]{title}{#1}
\renewbibmacro{in:}{%
  \ifentrytype{article}{}{\printtext{\bibstring{in}\intitlepunct}}}
\AtEveryBibitem{%
  \iffieldundef{eprint}{}{%
    \clearfield{journaltitle}%
    \clearfield{volume}%
    \clearfield{number}%
    \clearfield{pages}%
    \clearfield{eid}%
  }%
}
\renewbibmacro*{issue+date}{%
  \ifentrytype{article}
    {}%
    {\printtext[parens]{\printfield{issue}\setunit*{\addspace}\usebibmacro{date}}}%
  \newunit}
\renewbibmacro*{organization+location+date}{%
  \iffieldundef{eprint}
    {\printlist{location}%
     \iflistundef{organization}
       {\setunit*{\addcomma\space}}
       {\setunit*{\addcolon\space}}%
     \printlist{organization}%
     \setunit*{\addcomma\space}%
     \usebibmacro{date}%
     \newunit}
    {}}
\renewbibmacro*{publisher+location+date}{%
  \printlist{location}%
  \iflistundef{publisher}
    {\setunit*{\addcomma\space}}
    {\setunit*{\addcolon\space}}%
  \printlist{publisher}%
  \ifentrytype{inproceedings}{}{\setunit*{\addcomma\space}\usebibmacro{date}}%
  \newunit}
\renewbibmacro*{addendum+pubstate}{%
  \printfield{addendum}%
  \newunit\newblock
  \printfield{pubstate}%
  \ifboolexpr{
    test {\ifentrytype{article}} or
    test {\ifentrytype{inproceedings}} or
    not test {\iffieldundef{eprint}}
  }
    {\setunit{\addcomma\space}\usebibmacro{date}}
    {}}
\DeclareFieldFormat{eprint}{arXiv\addcolon\space\textit{#1}}
\renewbibmacro*{eprint}{\printfield{eprint}}

\usepackage[capitalize,noabbrev]{cleveref}

\newcommand{\citet}{\textcite}

\title{Depth-Recurrent Attention Mixture (Dreamer): \\ Giving Latent Reasoning the Attention it Deserves}

\author{%
    Jonas Knupp$^{1,2,3,*}$
    \And
    Jan Hendrik Metzen$^{1}$
    \And
    Jeremias Bohn$^{4}$
    \AND
    Georg Groh$^{4}$
    \And
    Kristian Kersting$^{2,5,6,7,8}$
    \AND \\[-1em]
    $^1$Aleph Alpha Research,
    $^2$Lab1141, \\
    $^3$Technical University of Munich (TUM), \\
    $^4$Research Group Social Computing, TUM, \\
    $^5$Computer Science Department, TU Darmstadt, \\
    $^6$Centre for Cognitive Science, TU Darmstadt, \\
    $^7$Hessian Center for Artificial Intelligence (hessian.AI), \\
    $^8$German Research Center for Artificial Intelligence (DFKI) \\[0.5em]
    $^*$\texttt{jonas.knupp@aleph-alpha-research.com}
}

\hypersetup{
    pdftitle={Depth-Recurrent Attention Mixture (Dreamer): Giving Latent Reasoning the Attention it Deserves},
    pdfauthor={Jonas Knupp, Jan Hendrik Metzen, Jeremias Bohn, Georg Groh, Kristian Kersting},
    pdfcreationdate={D:20260129114438Z},
    pdfmoddate={D:20260129114438Z},
    pdfdate={D:20260129114438Z},
    pdfmetadate={D:20260129114438Z},
    keeppdfinfo,
}

\begin{document}

\maketitle

\begin{abstract}
Depth-recurrence promises improved latent reasoning by sharing parameters across depths.
However, prior work still relies on partially fixed layer stacks and overlooks the bottleneck of constant hidden-size; moreover, it largely lacks rigorously resource-matched ablations.
To address this, we introduce depth-recurrent attention mixtures (Dreamer), a modular architecture with a single recurring layer.
Concretely, we mix sequence attention, depth attention, and sparse expert attention. This alleviates the hidden-size bottleneck, decouples scaling dimensions, and achieves effective yet efficient depth-recurrence.
Across natural language reasoning benchmarks at up to ${\sim}2$B parameters, Dreamer requires on average ${\sim}3\times$ fewer training tokens for the same accuracy as FLOP-, parameter-, and memory-matched modern Transformers, and outperforms ${\sim}2\times$ larger Transformers given same training tokens.
Analysis further reveals $2\text{--}11\times$ greater expert selection diversity than conventional MoEs, highlighting the flexible knowledge sharing across depths.
\end{abstract}

\vfill
\begin{figure}[H]%
    \centering%
    \includegraphics[width=\linewidth]{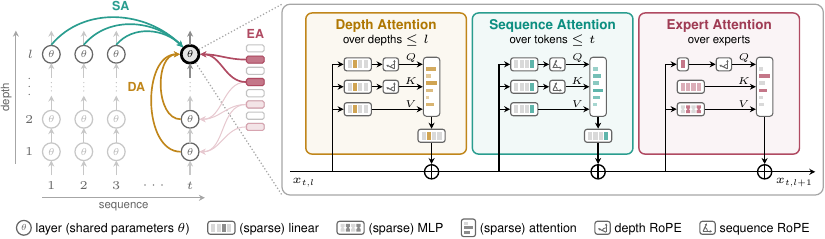}%
    \caption{%
        \textbf{D}epth-\textbf{re}current \textbf{a}ttention \textbf{m}ixtu\textbf{re} (\textbf{Dreamer}).
        This instance combines attention over sequence (horizontal), depth (vertical), and sparse experts (right) in a single recurring layer, to \textbf{facilitate knowledge reuse, alleviate hidden- and layer-size bottlenecks, improve reasoning and data efficiency, and decouple scaling dimensions} (params, FLOPs, depth, and depth-wise memory).
    }%
    \label{fig:dreamer-illustration-unfolded}%
\end{figure}

\clearpage

\section{Introduction}
\label{sec:intro}
Reasoning language models are increasingly central to modern AI applications, solving problems via chain-of-thought (CoT) \cite{weiChainofThoughtPromptingElicits2023,liChainThoughtEmpowers2024}.
However, traditional CoT performs recurrent reasoning only in the sequence dimension, while the depth dimension remains statically layered.
This limits knowledge reuse and compositional generalization, a crucial property for out-of-distribution performance.
Moreover, it enforces discrete verbalization, restricting expressivity and causing long, compute-expensive sequences compared to continuous latent space \cite{zhu2025reasoning,hao2024training}.

A potential solution is latent reasoning\footnote{
    We use the term ``latent reasoning'' to denote iterative computation in latent space without generating intermediate textual reasoning traces. We do not mechanistically establish which internal computations constitute a particular form of reasoning.
} via depth recurrence (DR) \cite{dehghani2018universal}, which eliminates static depth-wise knowledge separation by reusing parameters across depths. This facilitates compositional generalization and naturally unsupervised, continuous CoT beyond natural language \cite{zhu2025reasoning,hao2024training}.

However, prior work on DR \cite{dehghani2018universal,baiParameterEfficientConformersSharing2022a,tan2023sparse,csordas2024moeut,baeRelaxedRecursiveTransformers2025,saunshiReasoningLatentThoughts2025a,geiping2025scaling,koishekenov2025encode,wu2025parallel,zhu2025scaling,mcleish2025teaching,darlow2025continuous,wang2025hierarchical,jolicoeur2025less} largely lacks joint, tightly resource-matched comparisons, leaving the question open: How can DR be beneficial even with tightly matched FLOPs, parameter counts, and memory usage?
And since prior work on DR, especially for natural language reasoning, often still relies on partially fixed layer stacks (underutilizing DR for knowledge reuse, homogeneity, interpretability, and more), we further ask: How can DR be effective for natural language reasoning even with a single recurring layer?

We start by identifying and addressing two bottlenecks for DR in large language models (LLMs):

\textbf{Layer-size bottleneck $\rightarrow$ expert attention}:
Scaling parameter counts of dense models is much more inefficient with DR, since the whole model is executed at each depth. To solve this, we decouple compute from parameters via a sparse mixture of experts (MoE). It allows scaling a single DR layer to the total size of a traditional model while roughly maintaining FLOPs, global knowledge capacity, and expert granularity. Crucially, in such settings, expert routing requires extra care, since (i) there are more experts per depth for same granularity, making routing more inefficient, and (ii) routing parameters are shared across depths, making depth-dependent routing more challenging. We address both by modeling MoEs as attention over experts, which includes low-rank, depth-position-encoded scoring via queries and keys. This yields our expert attention (EA) (\cref{sec:ea}).

\textbf{Hidden-size bottleneck $\rightarrow$ depth attention}:
Pushing more reasoning from sequence into depth dimension exacerbates the hidden-size bottleneck, which imposes a constant memory size on latent depth-wise reasoning. Attention already solves the corresponding issue observed in recurrent neural networks (RNNs) along the sequence dimension, leading to sequence attention (SA). We suggest to leverage the same principle along the depth dimension, leading to depth attention (DA) (\cref{sec:da}).

We arrive at a unified view of SA, DA, and EA as an attention mixture over sequence, depth, and experts (\cref{fig:dreamer-illustration-unfolded,sec:methods}).
This provides a unification of knowledge access along all major dimensions via attention.
Such homogeneous approach may help to better understand and control model behavior, and facilitates modular extensibility via additional attention dimensions (\cref{sec:discussion}).

In summary, our contributions include:

    \textbf{Methods:} We introduce depth-recurrent attention mixtures (\cref{fig:dreamer-illustration-unfolded,sec:methods}), a modular, unifying framework that combines, e.g., sequence attention (\cref{sec:sa}), (optional) depth attention (\cref{sec:da}), and sparse expert attention (\cref{sec:ea}).
    Crucially, it adapts sparse MoEs for depth recurrence and makes DR efficient and effective.

    \textbf{Experiments:} We perform tight, joint FLOP-, parameter-, memory-, and data-matching (\cref{sec:resource_matching}) to compare non-DR (modern Transformers), DR (Dreamer), and various ablations in-between (\cref{sec:reasoning,sec:knowledge_and_language}).
    We show that even in this rigorously resource-matched regime and with a single recurring layer, appropriately designed DR models like Dreamer can yield substantial accuracy and efficiency gains for natural language reasoning. With $\sim$13\% inference slowdown in an unoptimized implementation, it achieves ${\sim}3\times$ better data efficiency and ${\sim}2\times$ better parameter\nobreakdash-, FLOP-, and memory-efficiency than modern Transformers.

    \textbf{Analysis:}
    We provide insights into knowledge usage across depths, from the perspectives of depth attention and expert attention (\cref{sec:analysis}).
    In high resolution (with over thousand small experts), we quantify knowledge capacity allocation and reuse patterns across depths, and derive practical suggestions for model design, for both DR and non-DR architectures.

\section{Related Work}

\textbf{Depth recurrence and parameter sharing across depth:}
Depth-recurrent (DR) Transformers are introduced by \citet{dehghani2018universal}.
While laying the groundwork, they miss resource-matched tests and use a dense DR core, making scaling intractable.
To address the scalability bottleneck of dense DR, \citet{tan2023sparse} investigate DR with sparse MoEs.
However, they still lack resource-matching and testing of natural language reasoning.
\citet{csordas2024moeut} perform more rigorous and scaled experiments with sparse DR, showing slight benefits of DR. However, they lack sparse baselines for fair comparisons. Moreover, they rely on multi-layer DR cores, in contrast to Dreamer's minimal, modular single-layer architecture.
\citet{baiParameterEfficientConformersSharing2022a,tanReXMoEReusingExperts2025b,li2025megrez2} investigate the reuse of MoE experts across (some adjacent) depths, thus resembling a form of partial/block-wise DR with MoEs. However, this prevents full depth-generalization.
The same limitation applies to \citet{baeRelaxedRecursiveTransformers2025}, exploring DR with depth-specific LoRA-adapters as relaxation.
In 2025, there was a surge of work on DR, for natural language reasoning \cite{saunshiReasoningLatentThoughts2025a,geiping2025scaling,koishekenov2025encode,wu2025parallel,zhu2025scaling,mcleish2025teaching} and small models for symbolic problems \cite{darlow2025continuous,wang2025hierarchical,jolicoeur2025less}.
Despite the intractability of scaling dense DR, most of these works use dense models.
Moreover, as proposed by \citet{geiping2025scaling}, many of them, especially for natural language, use dedicated encoders/decoders around the DR core.
In contrast, we show that purely a single-layer DR core is sufficient. It can learn such dedicated roles of encoding/decoding on its own if useful, while maintaining a homogeneous architecture, maximizing knowledge reuse and facilitating interpretability and modular extensibility. Further, it allows batching of inference steps in different depths, e.g.\ for speculative decoding and/or batched sequences with different/dynamic depths.
In addition, all mentioned works on DR so far ignore the hidden-size bottleneck, which becomes more pronounced as depth increases and more reasoning shifts from the sequence into the depth dimension. In contrast, our work addresses both hidden-size bottleneck and dense layer-size bottleneck.

\textbf{Other latent reasoning approaches:}
\citet{jaeglePerceiverGeneralPerception2021a} decouple input and latent sequence dimensions by cross-attending to the input sequence from a latent, recurrent dimension. However, training is not naively parallelizable across the output sequence.
\citet{hao2024training} generate continuous latent tokens between discrete I/O tokens in the sequence dimension. However, this neither supports naive sequence-parallel training nor leverages knowledge-reuse across depths, and latent steps extend the sequence length, increasing memory and compute requirements.
In contrast, our DR with depth attention operates on the depth dimension, overwriting depth-attention keys/values after each token, thus incurring only small compute and memory overhead while supporting sequence-parallel training.

\textbf{Depth attention and skip connections:}
As precursor to depth attention (DA), \citet{huang2017densely} explore skip connections between all layers, although only for shallow convolution networks for vision.
A different approach by \citet{pagliardiniDenseFormerEnhancingInformation2024a} averages hidden-states from previous layers, weighted by fixed learned weights.
\citet{xiao2025muddformer} compute dynamic weights for such aggregation via an MLP.
All these methods still do not resemble proper dot-product attention along the depth.
In contrast, \citet{fangCrosslayerRetrospectiveRetrieving2023a,clasterAdaptiveIntegratedLayered2025a} use dot-product DA; however, only for small vision models, time-series, or sentiment analysis, and neither natural language nor depth recurrence.
Moreover, these works do not tightly FLOP-, parameter-, and memory-match to compensate for the overhead of DA.
Further, the effect of DA in the context of depth recurrence is still unexplored.
This aspect is especially of interest because DA is designed to address the hidden-size bottleneck, which may be even more beneficial for depth-recurrent models, since they are intended to scale in depth to support more complex latent reasoning.

\section{Methods}\label{sec:methods}

\begin{wrapfigure}{r}{0.33\textwidth}%
    \vspace{-1.346em}%
    \centering
    \includegraphics{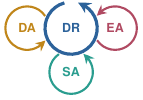}%
    \caption{
        Dreamer, as high-level folded view,
        combining depth recurrence (DR)
        with depth attention (DA), sequence attention (SA), and expert attention (EA).
    }%
    \label{fig:dreamer-folded}%
    \vspace{-1.5em}%
\end{wrapfigure}
We construct a layer that combines attention over multiple dimensions.
While such mixture of attention is proposed as general concept, the instantiation tested in this work comprises three specific orthogonal attention dimensions: sequence, depth, and experts  (\cref{fig:dreamer-folded}).
In our case, (only) the expert attention is sparse.
The layer is applied repeatedly, for latent recurrent computation, knowledge sharing across depths, and thus compositional generalization.
\Cref{fig:dreamer-illustration-unfolded} illustrates the architecture, with vertical attention over the depth, horizontal attention over the sequence, and sparse attention over MLP experts shared across depths.

\subsection{Attention (Background)}
\label{sec:attn}

Single-head self-attention of Transformers \cite{vaswaniAttentionAllYou2017b}, with sequence length $m$, key size $d_k$, value size $d_v$, aggregates values $V$ $[m, d_v]$ via scaled dot-product between queries $Q$ $[m, d_k]$ and keys $K$ $[m, d_k]$:
\begin{equation}
    \begin{aligned}
        & \operatorname{Attn}(Q, K, V) = \sigma\left(\frac{QK^T}{\sqrt{d_k}}\right) V
    \end{aligned}
    \label{eq:attn}
\end{equation}
By default, the activation function $\sigma$ is defined as $\operatorname{softmax}$ with upper triangular mask for causal attention.
The generalization to multi-head \cite{vaswaniAttentionAllYou2017b} and grouped-query attention \cite{ainslieGQATrainingGeneralized2023b} applies as usual.

\subsection{Sequence Attention (SA)}
\label{sec:sa}

The standard Transformer-like instance of $\operatorname{Attn}$ operates along the sequence dimension (i.e.\ $\operatorname{Attn}$'s $m \equiv$ sequence length). We therefore call it sequence attention (SA), to disambiguate our attention variants.
$Q$, $K$, and $V$ are projected from normalized hidden-states. In our case, $Q$ and $K$ are position-encoded by RoPE \cite{suRoformerEnhancedTransformer2024} and normalized by RMSNorm \cite{zhangRootMeanSquare2019b}.

\subsection{Depth Attention (DA)}
\label{sec:da}

\begin{algorithm}[H]
    \caption{Depth Attention}
    \label{alg:da}
    \begin{Verbatim}[commandchars=\\\{\}]
b, s, h = x.shape      \textcolor[HTML]{888888}{# Batch, sequence, hidden dimension}
x = x.view(b*s, 1, h)  \textcolor[HTML]{888888}{# Merge sequence into batch, create depth dimension}
y = attn(x, kv_cache)  \textcolor[HTML]{888888}{# Attention as usual over second dimension}
y = y.view(b, s, h)    \textcolor[HTML]{888888}{# Recover sequence dimension}
    \end{Verbatim}
\end{algorithm}
Analogously to SA, depth attention (DA) applies a separate instance of attention over the previous depths of the same token ($\operatorname{Attn}$'s $m \equiv$ depth).
This can be seen as a transposed/vertical Transformer.
Similar to how SA in standard Transformers addresses the hidden-size bottleneck of RNNs in the sequence direction, DA addresses it in the depth direction.
Moreover, in DA, RoPE encodes the depth instead of sequence positions. We refer to this depth-position encoding as depth-RoPE.

We implement DA by treating the sequence as batch and depth as sequence dimension. This allows leveraging existing optimized kernels.
A basic PyTorch implementation is given in \cref{alg:da}.
Since DA is inherently sequential, we use a (differentiable, in-place) kv-cache even during training.

KV-caching works as usual. But contrary to SA, we can drop/overwrite the cache for DA after each token during inference, since DA operates on a per-token cache across depths while SA operates on a per-depth cache across tokens. Hence, the memory overhead of DA is constant wrt.\ the sequence length and negligible compared to the usually much larger cache for SA (\cref{fig:efficiency}).

\subsection{Expert Attention (EA)}
\label{sec:ea}

We instantiate a third attention variant, now operating over submodels called experts ($\operatorname{Attn}$'s $m \equiv$ number of experts), hence the name expert attention (EA).
This resembles a special kind of mixture of experts (MoE), carrying the majority of static model knowledge, to be now also accessed via attention.
Experts (usually multi-layer perceptrons, MLPs) compute values $V_{\operatorname{EA}}$ $[m, d_v]$.
Keys $K_{\operatorname{EA}}$ $[m, d_k]$ are provided as learnable parameters per expert, and a raw query $Q_{\operatorname{EA}}$ $[1, d_k]$ is computed by projecting hidden-state $x$ $[h]$ into the $d_k$-dimensional key space, as usual for attention.
For simplicity, we use single-head attention for EA here.

Unlike standard MoEs, EA computes routing score logits similar to DA, retaining benefits of attention:
For depth-aware expert selection, we apply depth-RoPE to queries, but with reversed depth for half of its dimensions, i.e.\ starting at the maximum depth and decreasing with depth.
This adds forward-looking information about the remaining number of depth-iterations.
Moreover, instead of the common computation of routing scores as $\sigma(W x)$ via a projection $W$ $[m, h]$, final attention scores are computed low-rank (key size $d_k$ $\ll$ hidden size $h$) via dot-product between queries and keys as in \cref{sec:attn}.
This allows scaling the number of experts more efficiently:
\begin{equation}
    \begin{aligned}
        & \operatorname{EA}(x) = \operatorname{Attn}(Q_{\operatorname{EA}}, K_{\operatorname{EA}}, V_{\operatorname{EA}}), \\
        & \text{with } V_{\operatorname{EA},e} = \operatorname{MLP}_e(x) \in \mathbb{R}^{d_v} \text{ for expert } e \in [m].
    \end{aligned}
\end{equation}

For further efficiency, we use top-$k$ sparse attention scores, similar to DeepSeek-V3 \cite{liu2024deepseek}, to execute only a subset of experts.
Accordingly, attention activations ($\sigma$ in \cref{eq:attn}) of EA are defined as
\begin{equation}\label{eq:ea-selection}
    \begin{aligned}
        & \sigma(x) = \operatorname{norm}(\operatorname{TopK}(\operatorname{sigmoid}(x), b, k)) \\
        & \text{with } \operatorname{TopK}(x, b, k)_{i,j} = x_{i,j} \cdot 1_{j \in \operatorname{TopKIds}(x_i + b, k)}.
    \end{aligned}
\end{equation}
Since sparsity can degenerate expert balancing, we balance via bias vector $b$, updated according to
\begin{equation}\label{eq:ea-balancing}
    \begin{aligned}
        & b \leftarrow b + \lambda \operatorname{sign}(\operatorname{median}(N) - N)
    \end{aligned}
\end{equation}
with update rate $\lambda$. $N_i$ counts how often expert $i$ was used since the last update.
This is slightly different from DeepSeek-V3.
We use no distributed expert-parallelism, thus we have no notion of expert overload.
To determine the bias update direction, we use the median expert usage count as baseline. Therefore, the statistical average of $b$ remains zero, which stabilizes long-term training.

\subsection{Attention Mixture}
\label{sec:attn_mixture}

Combining DA, SA, and EA can be done sequentially:
\begin{equation}\label{eq:attn_mix_seq}
    \begin{aligned}
        & y_{\operatorname{DA},l} = x_l + \operatorname{DA}(x_{:l}),
            \quad y_{\operatorname{SA},l} = y_{\operatorname{DA},l} + \operatorname{SA}(y_{\operatorname{DA},l}),
            \quad y_l = y_{\operatorname{SA},l} + \operatorname{EA}(y_{\operatorname{SA},l})
    \end{aligned}
\end{equation}
In such a declarative form, at each depth $l$, DA consumes all previous hidden-states $x_{:l}$ $[l, b, s, h]$ with batch size $b$, sequence length $s$, and hidden-size $h$, while the other components and outputs are only concerned with the latest depth's states.
In practice however, only the latest hidden state is passed on, together with cached keys/values for DA and SA.
Regarding the order of DA, SA, and EA, we saw only small accuracy differences, especially with large depth.

For improved throughput, DA, SA, and/or EA may be parallelized, either partially (e.g. $y_{\operatorname{DA+SA},l} = x_l + \operatorname{DA}(x_{:l}) + \operatorname{SA}(x_l),\ y_l = y_{\operatorname{DA+SA},l} + \operatorname{EA}_l(y_{\operatorname{DA+SA},l})$) or fully ($y_l = x_l + \operatorname{DA}(x_{:l}) + \operatorname{SA}(x_l) + \operatorname{EA}(x_l)$).
In tests with $\sim$1B parameters, each additional parallelization yielded $\sim$15\% speedup at the cost of up to 10--20\% higher benchmark error rates.
For this work, we follow \cref{eq:attn_mix_seq}.

The concept of attention mixtures as general foundation is modular and extensible. Depending on the application, different attention dimensions may be added. Some options are mentioned in \cref{sec:discussion}.

\subsection{Depth Recurrence (DR)}
\label{sec:dr}

Reusing layers/weights across depths seems straightforward. However, when increasing the depth, the scales of residuals and their additive updates can become incompatible across depths, which prevents learning depth-generalized experts. Hence, for our depth-recurrent variants, we apply RMSNorm to the residual stream itself, obtaining the layer's next input $x_{l+1} = \operatorname{Norm}(y_l)$.

As another intricacy of depth recurrence, enforcing the attention modules to reuse their projection weights across depths leads to worse performance (\cref{tab:reasoning-math-ablation-results}).
To mitigate this, while maintaining a fully depth-recurrent architecture, we replace the fused QKV input projection and the multi-head output aggregation with novel linear MoEs, each with $E$ separate experts, $W_1, \ldots, W_E$.
Routing and balancing follow \cref{eq:ea-selection,eq:ea-balancing}, but with linear experts, top-1 selection ($k{=}1$), and no score normalization.
Routing is tied across both projections, reducing overhead and stabilizing training.

In top-1 MoEs, learning good routing scores becomes harder, because all information about alternative experts is removed from the gradient.
We address this with a shared, always-active expert $W_{\text{shared}}$, serving as a stable anchor for routing score learning and balancing.
Contrary to usual shared experts in MoEs, we scale it by the active routing score, with stop-gradient $\operatorname{sg}(\cdot)$.
Letting $\sigma \in \mathbb{R}^E$ be the routing score vector (following \cref{eq:ea-selection}, $k{=}1$, no normalization) and $V = [xW_1; \ldots; xW_E]$ be the stacked expert outputs, each linear MoE produces
\begin{equation}\label{eq:attn-moe}
    y_{\text{moe}} = \sigma V + \operatorname{sg}(\max \sigma)\, x W_{\text{shared}}.
\end{equation}
Linear experts and equal scaling of shared and routed experts allows \cref{eq:attn-moe} to be reduced to $\max(\sigma)\, x (W_{\operatorname{argmax} \sigma} + W_{\text{shared}})$ at inference.
Hence, $W_{\text{shared}}$ can be absorbed into each $W_e$ prior to inference, eliminating its compute and memory overhead entirely.
Compared to standard attention, only the MoE routing remains as slight overhead, which we account for in our FLOP-matching.

\subsection{Resource-Matched Architecture Comparison}\label{sec:resource_matching}

Comparisons between DR and non-DR often suffer from confounding factors, like non-matching parameter counts, memory requirements, and numbers of floating-point operations (FLOPs) per token.
To address this, we perform rigorous resource-matching across all models of the same depth. For this, we require a shared base architecture across all variants, and decoupling of FLOPs from parameters via sparse MoEs. To match sequential FLOPs for both training and inference, we deliberately use fixed-depth DR, which also avoids conflation of DR and dynamic depth.

We compare multiple Dreamer variants: with DA, without DA ($-$DA), other ablations, as well as a non-DR variant, resembling modern Transformers like Qwen3 \cite{yangQwen3TechnicalReport2025} and DeepSeek-V3 \cite{liu2024deepseek} but with stable MoE balancing like Dreamer. This minimizes confounding factors, for fair comparisons and isolating effects of DR, DA, and other design decisions. Hyperparameters are in \cref{tab:hparams}.

For tight and joint resource matching, we use bivariate coordinate descent, alternating between FLOP and parameter matching:
FLOPs are matched by adjusting EA MLP intermediate sizes, and parameters are matched by adjusting the EA expert counts, both via binary search.
Changing the expert count slightly affects EA routing FLOPs, hence we repeat the alternating loop until convergence (usually 2 iterations).
Thus, the slight compute overheads of DA and DR's larger EA router are compensated fairly, for meaningful comparison with very similar resource requirements (\cref{tab:model_resource_requirements}).
Since there is no degree of freedom left to also align speed, there remains an inference slowdown of $\sim$13\% with DA due to memory movements between HBM and the on-chip cache hierarchy.
\Cref{sec:discussion} discusses some solutions.
\begin{table}[t]
    \footnotesize
    \caption{Model resource requirements. FLOPs per token and peak memory are measured over 1024 generated tokens. Parameters exclude vocab. As intended, our matching closely aligns resources.}
    \label{tab:model_resource_requirements}
    \begin{center}
    \begin{small}
    \setlength{\tabcolsep}{3.5pt}
    \begin{tabular}{lcccccccccccc}
        \toprule
        & \multicolumn{4}{c}{\textsc{\textbf{Parameters (B)}}} & \multicolumn{4}{c}{\textsc{\textbf{FLOPs (B/token)}}} & \multicolumn{4}{c}{\textsc{\textbf{Memory (GB)}}} \\
        \cmidrule(lr){2-5} \cmidrule(lr){6-9} \cmidrule(lr){10-13}
        \textsc{\textbf{Depth $\triangleright$}} & 8 & 16 & 24 & 32 & 8 & 16 & 24 & 32 & 8 & 16 & 24 & 32 \\
        \midrule
        Transformer & 0.454 & 0.908 & 1.362 & 1.816 & 0.601 & 0.939 & 1.277 & 1.615 & 1.511 & 2.452 & 3.394 & 4.336 \\
        Dreamer $-$DA & 0.455 & 0.908 & 1.362 & 1.817 & 0.600 & 0.939 & 1.279 & 1.620 & 1.512 & 2.460 & 3.411 & 4.379 \\
        Dreamer & 0.454 & 0.908 & 1.363 & 1.816 & 0.601 & 0.941 & 1.273 & 1.613 & 1.511 & 2.461 & 3.414 & 4.380 \\
        \bottomrule
    \end{tabular}
    \end{small}
    \end{center}
\end{table}

\begin{figure}[H]
    \centering
    \centerline{\includegraphics[width=\linewidth]{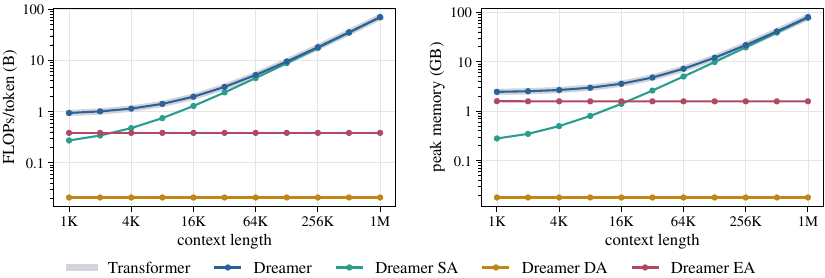}}%
    \caption{Inference FLOPs per token and peak memory (log-scaled) for Dreamer and Transformer (depth 16) over context lengths, decomposed into Dreamer's per-component contributions (SA, DA, EA). Resource requirements for Dreamer and Transformer stay tightly matched across context lengths, with SA dominating. DA memory usage (18 MB) stays constant due to per-token kv-cache eviction.}
    \label{fig:efficiency}
\end{figure}

\Cref{fig:efficiency,tab:efficiency-flops,tab:efficiency-memory} show inference compute and memory usage for Dreamer and Transformer at varying context lengths up to 1M, in total and per Dreamer component (SA, DA, EA).

In terms of complexity, for context length $L$, DA and EA require a total compute of $O(L)$ and total memory of $O(1)$ while SA requires a total compute of $O(L^2)$ and total memory of $O(L)$. Since SA is equal-shaped and strictly dominates for Dreamer and standard Transformer, at same depth the resource consumptions converge to SA's usage as context length increases.

\section{Experiments}
\label{sec:experiments}

In the following, we first describe our experimental setup. We then discuss evaluation results and further investigate the latent behavior of our depth-recurrent models through the lens of DA and EA.

\subsection{Reasoning Language Modeling}
\label{sec:reasoning}

We test natural language and math reasoning via common benchmarks and educational math text generation (\cref{tab:reasoning-math-benchmarks}), jointly covering logical reasoning, factual knowledge, language understanding and generation.
We train on $\sim$100B tokens of instruction datasets (\cref{tab:reasoning-math-datasets}), cleaned, deduplicated, and decontaminated.
We avoid artificial CoT traces, due to their length and low signal-to-noise ratio, and since we target latent reasoning.
Our training data still contain concise solution explanations, balancing intermediate guidance and reduced verbalization, relying on latent computations for details.

\Cref{tab:reasoning-math-results} shows a consistent, strong benefit of Dreamer over modern Transformers across all tests, and ablations in \cref{tab:reasoning-math-ablation-results} show that all design decisions, and in particular EA, improve accuracy.
DA helps modestly despite the rigorous FLOP-matching.
We expect future work to further improve this, e.g.\ via more efficient DA variants, as we will discuss in \cref{sec:discussion}.
In total, Dreamer's data efficiency gains average to 3.3$\times$.
Accuracy gains average to 10.1 percentage points (ppt).
Moreover, Dreamers on average outperform modern Transformers of twice the depth. Given the resource metrics in \cref{tab:model_resource_requirements}, this yields a ${\sim}2\times$ reduction in parameters, FLOPs, and peak memory.

\begin{table}[t]
    \footnotesize
    \caption{%
        Natural language reasoning \& math accuracies (Acc), validation perplexity (PPL), and data efficiency (DE). DE is the saving factor of training tokens to reach Transformer's best accuracy.
        All models are pre-trained on $\sim$100B tokens.
        Models of same depth are tightly FLOP-, parameter-, and memory-matched.
        Dreamer consistently outperforms modern Transformers, even without DA.
    }
    \label{tab:reasoning-math-results}
    \begin{center}
    \begin{small}
    \begin{tabular}{lccccccc}
        \toprule
        \textsc{\textbf{Model}} & \hspace{-0.6em}\textsc{\textbf{Depth}}\hspace{-0.6em} & \textsc{\textbf{Avg}} & \textsc{\textbf{MathQA}} & \textsc{\textbf{GSM8K}} & \textsc{\textbf{MATH}} & \textsc{\textbf{MMLU}} & \textsc{\textbf{Maths-}} \\
        & & & {\fontsize{8pt}{9pt}\selectfont (0-shot)} & {\fontsize{8pt}{9pt}\selectfont (0-shot)} & {\fontsize{8pt}{9pt}\selectfont (0-shot)} & {\fontsize{8pt}{9pt}\selectfont (math, 0-shot)} & \textsc{\textbf{College}} \\
        & & Acc/DE {\tiny$\uparrow$} & Acc/DE {\tiny$\uparrow$} & Acc/DE {\tiny$\uparrow$} & Acc/DE {\tiny$\uparrow$} & Acc/DE {\tiny$\uparrow$} & PPL {\tiny$\downarrow$} \\
        \midrule
        Transformer & 8 & 26.8 / 1.0 & 19.9 / 1.0 & 30.4 / 1.0 & 34.3 / 1.0 & 22.7 / 1.0 & 7.47 \\
        Dreamer $-$DA & 8 & \underline{34.6} / \underline{2.3} & \underline{29.9} / \textbf{\underline{3.6}} & \underline{35.3} / \underline{1.8} & \underline{37.5} / \underline{1.5} & \textbf{\underline{35.5}} / \underline{2.8} & \textbf{\underline{7.32}} \\
        Dreamer & 8 & \textbf{\underline{35.6}} / \textbf{\underline{2.8}} & \textbf{\underline{32.6}} / \underline{3.3} & \textbf{\underline{38.4}} / \textbf{\underline{2.5}} & \textbf{\underline{38.7}} / \textbf{\underline{2.3}} & \underline{34.0} / \textbf{\underline{3.4}} & \underline{7.36} \\
        \midrule
        Transformer & 16 & 32.3 / 1.0 & 22.9 / 1.0 & 43.8 / 1.0 & 40.9 / 1.0 & 26.6 / 1.0 & 6.74 \\
        Dreamer $-$DA & 16 & \textbf{\underline{45.7}} / \textbf{\underline{4.0}} & \textbf{\underline{45.6}} / \textbf{\underline{5.4}} & \underline{47.6} / \underline{1.6} & \underline{47.8} / \underline{2.8} & \textbf{\underline{44.5}} / \textbf{\underline{4.8}} & \underline{6.64} \\
        Dreamer & 16 & \underline{43.4} / \underline{2.9} & \underline{37.1} / \underline{3.2} & \textbf{\underline{51.7}} / \textbf{\underline{2.4}} & \textbf{\underline{50.7}} / \textbf{\underline{3.5}} & \underline{38.5} / \underline{3.4} & \textbf{\underline{6.41}} \\
        \midrule
        Transformer & 24 & 39.9 / 1.0 & 33.6 / 1.0 & 45.1 / 1.0 & 43.2 / 1.0 & 39.0 / 1.0 & 6.71 \\
        Dreamer $-$DA & 24 & \underline{46.7} / \underline{2.6} & \underline{40.7} / \textbf{\underline{3.2}} & \textbf{\underline{51.9}} / \underline{2.0} & \textbf{\underline{51.5}} / \textbf{\underline{3.5}} & \underline{43.2} / \underline{1.6} & \underline{6.55} \\
        Dreamer & 24 & \textbf{\underline{47.3}} / \textbf{\underline{2.7}} & \textbf{\underline{42.2}} / \underline{2.6} & \underline{51.8} / \textbf{\underline{2.3}} & \underline{49.3} / \underline{3.4} & \textbf{\underline{47.8}} / \textbf{\underline{2.8}} & \textbf{\underline{6.21}} \\
        \midrule
        Transformer & 32 & 39.5 / 1.0 & 26.0 / 1.0 & 49.7 / 1.0 & 48.2 / 1.0 & 37.2 / 1.0 & 6.31 \\
        Dreamer $-$DA & 32 & \underline{47.6} / \underline{2.6} & \underline{43.2} / \underline{5.4} & \underline{51.5} / \underline{1.5} & \underline{51.3} / \underline{2.1} & \underline{45.3} / \underline{1.9} & \underline{6.12} \\
        Dreamer & 32 & \textbf{\underline{51.4}} / \textbf{\underline{3.5}} & \textbf{\underline{50.8}} / \textbf{\underline{8.3}} & \textbf{\underline{56.3}} / \textbf{\underline{2.3}} & \textbf{\underline{54.5}} / \textbf{\underline{2.6}} & \textbf{\underline{50.0}} / \textbf{\underline{4.0}} & \textbf{\underline{5.90}} \\
        \bottomrule
    \end{tabular}
    \end{small}
    \end{center}
    \vskip -0.1in
\end{table}

\begin{table}[b]
    \footnotesize
    \caption{%
        Dreamer ablations.
        ``$-$revDPE'': no half-reversed depth-position encoding, instead standard RoPE.
        ``$-$AMoE'': no linear MoE for attention projections, instead standard attention projections.
        ``$-$DPE'': no depth-position for EA, instead pure low-rank EA.
        ``$-$EA'': no EA, instead modern MoE.
    }
    \label{tab:reasoning-math-ablation-results}
    \begin{center}
    \begin{small}
    \begin{tabular}{lccccc}
        \toprule
        \textsc{\textbf{Model}} & \textsc{\textbf{Avg}} & \textsc{\textbf{MathQA}} & \textsc{\textbf{GSM8K}} & \textsc{\textbf{MATH}} & \textsc{\textbf{MMLU}} \\
        {\fontsize{8pt}{9pt}\selectfont (depth 16)} & & {\fontsize{8pt}{9pt}\selectfont (0-shot)} & {\fontsize{8pt}{9pt}\selectfont (0-shot)} & {\fontsize{8pt}{9pt}\selectfont (0-shot)} & {\fontsize{8pt}{9pt}\selectfont (math, 0-shot)} \\
        & Acc {\tiny$\uparrow$} & Acc {\tiny$\uparrow$} & Acc {\tiny$\uparrow$} & Acc {\tiny$\uparrow$} & Acc {\tiny$\uparrow$} \\
        \midrule
        Transformer & 32.3 & 22.9 & 43.8 & 40.9 & 26.6 \\
        Dreamer $-$DA $-$EA & 33.2 & 24.4 & 45.5 & 47.4 & 35.3 \\
        Dreamer $-$DA $-$DPE & 33.5 & 31.7 & 40.9 & 43.4 & 29.4 \\
        Dreamer $-$DA $-$AMoE & 34.1 & 26.4 & 45.0 & 45.0 & 34.5 \\
        Dreamer $-$DA $-$revDPE & 42.9 & 40.2 & 45.1 & 47.2 & 40.7 \\
        Dreamer $-$DA & \textbf{45.7} & \textbf{45.6} & \textbf{47.6} & \textbf{47.8} & \textbf{44.5} \\
        \bottomrule
    \end{tabular}
    \end{small}
    \end{center}
    \vskip -0.1in
\end{table}

\subsection{Broad Knowledge and Language Modeling}
\label{sec:knowledge_and_language}

Our work focuses on reasoning. However, we also verify that Dreamer does not degrade on broader knowledge and language modeling tasks.
For this, we train on 100B tokens of diverse natural language data (\cref{tab:knowledge-language-datasets}) and test on standard benchmarks for language modeling and broad knowledge (\cref{tab:knowledge-language-benchmarks}).
\cref{tab:knowledge-language-results} shows no degradation on these benchmarks, instead a consistent benefit of Dreamer over Transformers across all tests.
The benefits are smaller than for reasoning, which is compatible with the intuition that DR mostly benefits reasoning due to its compositional, recursive nature.

\begin{table}[t]
    \footnotesize
    \caption{%
        Broad knowledge and language modeling benchmark results.
        We compare Dreamer to modern Transformer architectures.
        All models are pre-trained on ca.\ 100B tokens, have depth 32, and are tightly FLOP-, parameter-, and memory-matched.
    }
    \label{tab:knowledge-language-results}
    \begin{center}
    \begin{small}
    \begin{tabular}{lccccc}
        \toprule
        \textsc{\textbf{Model}} & \textsc{\textbf{Avg}} & \textsc{\textbf{HellaSwag}} & \textsc{\textbf{MMLU}} & \textsc{\textbf{PIQA}} & \textsc{\textbf{TriviaQA}} \\
        {\fontsize{8pt}{9pt}\selectfont (depth 32)} & Acc {\tiny$\uparrow$} & Acc {\tiny$\uparrow$} & Acc {\tiny$\uparrow$} & Acc {\tiny$\uparrow$} & Acc {\tiny$\uparrow$} \\
        \midrule
        Transformer & 34.5 & 38.1 & 29.3 & 69.0 & 1.6 \\
        Dreamer $-$DA & 36.2 & 39.6 & \textbf{30.6} & 69.5 & 5.1 \\
        Dreamer & \textbf{37.4} & \textbf{40.6} & \textbf{30.6} & \textbf{71.4} & \textbf{7.0} \\
        \bottomrule
    \end{tabular}
    \end{small}
    \end{center}
    \vskip -0.1in
\end{table}

\subsection{Analyzing Depth Attention and Expert Attention}
\label{sec:analysis}

\begin{wrapfigure}{r}{0.425\textwidth}%
    \vspace{-1.5em}%
    \centering
    \centerline{\includegraphics[width=\linewidth]{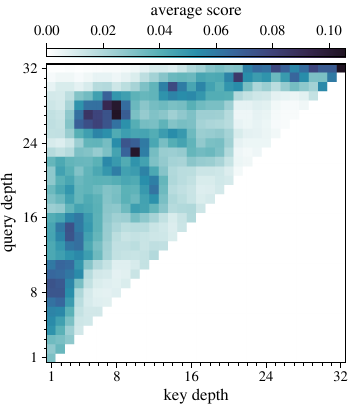}}%
    \caption{Map of average depth attention scores, normalized per depth.}
    \label{fig:da-map}
    \vspace{-0.9em}%
\end{wrapfigure}
We collect statistics of Dreamer (depth 32) on 1000 random sequences from validation data, to analyze the learned behavior of DA and EA.
\cref{fig:da-map} shows the average attention scores of DA.
This aggregate view reveals nontrivial patterns:
Overall, there are strong clusters in some depths, with clear phase transitions, as well as regions with more distributed attention.
Typically, early depths look back to the first few depths, followed by more diverse DA patterns, after which middle depths are recalled. In the end, middle to high depths inform the final output.
While further interpretability/explainability work is needed to properly understand such patterns, they already indicate that DA meaningfully extends the expressivity beyond uniform skip-connections, leading to more targeted information routing between depths.

\begin{figure}[t]
    \centering
    \centerline{\includegraphics[width=\linewidth]{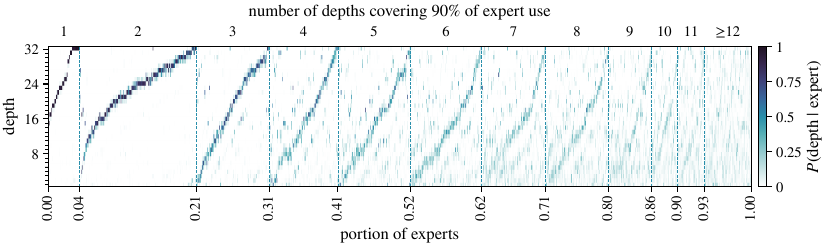}}%
    \caption{Distribution of depths per expert. Experts are sorted by the number of depths in their top 90th percentile of sorted $P(\text{depth}|\text{expert})$, as a measure of depth-generalization. Within these groups, experts are sorted by sampled depth from their distributions.
    Observations: Higher depths use more depth-specialized experts. Ca.\ 50\% of experts are widely depth-generalized.}
    \label{fig:ea-map-depth-prob}
\end{figure}

We investigate EA behavior by showing the distribution of depths per expert in \cref{fig:ea-map-depth-prob}.
Since experts have no inherent order, we group and sort them by the number of depths in the top 90th percentile of $P(\text{depth}|\text{expert})$, effectively sorting by increasing depth-generalization.
Within each group, we sort by a sample from their depth distributions, yielding a meaningful order while faithfully representing the distributions.
The resulting plot reveals several insights and implications:
The later half of depths (i.e.\ depths $\ge$ 50\% of maximum depth) use most of the highly depth-specific experts (i.e.\ with $P(\text{depth}|\text{expert}) > 0.5$). This results in $\sim$22\% of experts preferring one or two depths $\ge$50\%. In particular, the last depth owns the most single-depth experts. Hence, one may model this behavior via a decoding layer.
In contrast, depths $<$50\% use more depth-generalized than -specific experts. Almost no experts are dedicated to one or two depths $<$50\%. Hence, for these depths, the rigid layer separation of traditional models is most restrictive.
We further observe that most experts with only two or three depths distribute their usage across adjacent depths ($\pm1$). Hence, they are not fully depth-generalized.
A sliding window over experts, or hierarchical expert subset selection, could model this pattern while reducing routing costs.
Finally, $\sim$50\% of experts are each used in at least 6 depths ($\sim$19\% of depths). This indicates knowledge reuse and depth-generalization beyond just adjacent depths. Therefore, the model makes heavy use of DR's extended flexibility. Hence, to facilitate such behavior, at least a subset of experts should be selectable from all depths.

\begin{wrapfigure}{r}{0.5\textwidth}%
    \vspace{-0.309em}
    \centering
    \centerline{\includegraphics[width=\linewidth]{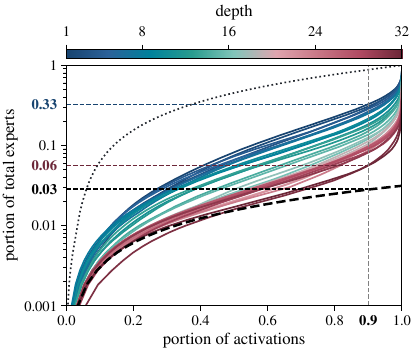}}%
    \caption{%
        Lorenz curves (log-scaled) for EA expert routing.
        This shows how many expert routings occurred vs.\ how many unique experts were used,
        for Dreamer per depth (colored), Dreamer globally (thin dotted), and normal MoEs per depth (= globally) (thick dashed).
        Observations: Dreamer uses 2--11$\times$ more experts per depth than normal MoEs. Global balance remains strong (Gini: 0.075).
    }
    \label{fig:ea-lorenz-curves}
\end{wrapfigure}
\Cref{fig:ea-lorenz-curves} relates each portion of expert activations (i.e.\ how many expert routings occurred) to the portion of total experts activated (i.e.\ how many experts were used). The plot therefore represents Lorenz curves (per depth and globally). A perfectly linear relationship (i.e.\ log-curve, due to log-scaled y-axis) means uniform sampling of experts.
We obtain the following observations and implications:
Low depths use $\sim$5$\times$ more experts than high depths. In particular, the last two depths use much fewer experts than all others.
Assuming that this is governed by the optimization objective and shows a generally favorable capacity allocation, it suggests that conventional MoEs' prevalent uniform distribution of experts across layers is suboptimal.
Further, Dreamer actively uses 2--11$\times$ more experts per depth than conventional MoEs. This large gain in flexibility of knowledge usage provides yet another argument for DR.
Finally, we see that our global expert balancing works well, given that the global Lorenz curve is almost linear, leaving no expert underutilized. Quantitatively, we obtain a low global Gini coefficient of 0.075.

\section{Discussion and Future Work}\label{sec:discussion}

\textbf{Limitations:}
Experiments are limited to fixed depth; dynamic depth is deliberately out-of-scope for reasons stated in \cref{sec:intro}.
The tested scale is moderate; results at larger scales and rigorous scaling laws remain to be investigated.
Our work focuses on reasoning; the broader-domain evaluation in \cref{sec:knowledge_and_language} is limited in scope.
The depth attention pattern analysis (\cref{sec:analysis}) is descriptive; establishing causal links between attention patterns and model behavior requires further investigation.

\textbf{Depth Attention:}
Since DA's kv-cache can be overwritten after each token, its memory footprint remains constant wrt.\ sequence length (\cref{fig:efficiency}), though not wrt.\ depth.
Compared to SA, which typically operates over much larger lengths, the compute overhead of DA is small, though not entirely negligible.
We want to stress that we compensate for this overhead through our very tight FLOP-matching.
Even in this rigorous setting, DA provides a nontrivial average accuracy gain.
Since deeper models are more bottlenecked by small hidden size, analogous to how RNNs are bottlenecked by hidden size at long sequences, depth attention may continue being beneficial at larger depths, but reliable scaling laws are still to be established.
\\
There remains a memory movement overhead, slowing down inference by $\sim$13\% with an unoptimized DA implementation. (Resource-matching covers FLOPs, parameters, and memory, leaving no degree of freedom for also tightly matching speed.)
We expect this to improve by future kernel optimizations and explorations of DA variations, e.g.\ sliding windows, dilation, linear attention variants, and modern RNNs/SSMs along the depth dimension. In general, we hope our contribution sparks more inspiration to treat depth as first-class sequential dimension, going beyond basic residual streams.

\textbf{Dynamic Depth:}
Our work deliberately focuses on fixed-depth DR for fair FLOP-matched comparisons and to avoid confounding evaluations (see \cref{sec:intro}).
Nonetheless, dynamic depth remains a natural extension for future work. Since our DA and EA use RoPE as depth position encoding, one may explore dynamic RoPE scaling in this context. Alternatives to RoPE may also be worth considering, possibly improving depth generalization. We believe that reliably generalizing DR beyond the trained depth regime is a critical and still underexplored problem.

\textbf{Attention Mixture:}
We unified information flows as attention along all major dimensions: sequence (like traditional attention), depth (like dynamic skip-connections), and experts (like MoEs).
This perspective may help understanding and controlling model behavior.
Moreover, while our work focuses on these three dimensions, attention mixtures are a generic, modular framework. For instance, sparse EA may be extended or replaced by parameter attention \cite{wangTokenFormerRethinkingTransformer2025}, and additional knowledge (learned latents or text embeddings) may be plugged in as separate attention dimension, resembling ``in-depth'' latent RAG.
As the number of attention dimensions increases, the described parallel attention mixture variant may then be preferable for efficiency, and components may be executed asynchronously or conditionally, similar to a higher-order sparse MoE. However, it remains an open question whether such approaches can be beneficial in practice, given sparsity-averse GPUs/TPUs/etc.

\section{Conclusion}

This work introduced Dreamer, a fully depth-recurrent (DR) attention mixture
with a single recurring layer that repeatedly attends over sequence, depth, and experts.
Intuitively, Dreamer targets two bottlenecks of DR: Sparse expert attention addresses the layer-size bottleneck and depth attention addresses the hidden-size bottleneck, together making DR efficient and effective.

We performed tightly FLOP-, parameter-, memory-, and data-matched comparisons between modern Transformers, Dreamer, and various variants in-between.
Our results show that even in a rigorously resource-matched regime and with a single recurring layer, DR can yield substantial accuracy and efficiency gains for natural language reasoning, when using advancements like expert attention.
Ultimately, Dreamer achieves ${\sim}3\times$ better data efficiency and ${\sim}2\times$ better parameter-, FLOP-, and memory-efficiency, at the cost of $\sim$13\% inference slowdown in an unoptimized implementation.
Our ablations highlight the importance of DR-aware architecture design like our expert attention.

Future work may further find inspiration in treating depth as first-class sequential dimension, going beyond the prevalent static layering and residual connections, exploring alternatives to vanilla depth attention, and adding other attention dimensions.

\section{Acknowledgments}
We thank Letiția Pârcălăbescu for valuable feedback on the paper draft.
We thank the Aleph Alpha infrastructure team for their work and support throughout this project.
We thank Tobias Ribizel for infrastructure support and the Chair of Computational Mathematics at TUM for compute resources during our early experiments for this project.
We thank Vladimir Golkov and Daniel Cremers for their personal support and for enabling various extracurricular research, which inspired this project.

\newpage
\printbibliography

\newpage
\appendix
\section{Appendix}

\subsection{Broader Impact Statement}
\label{sec:broader_impact}
This work enhances the efficiency and reasoning capabilities of AI by introducing a modular, depth-recurrent attention mixture framework. By enabling ${\sim}3\times$ better data efficiency and ${\sim}2\times$ better parameter- and FLOP-efficiency, we make it easier to build high-performance models, contributing to making AI potentially more sustainable. This allows complex reasoning tasks to be performed with significantly fewer computational resources, which is essential for deploying advanced AI in the wild. However, we acknowledge potential risks. Because our methods improve the model's ability to reason internally (i.e.\ latent reasoning), it may become harder for humans to interpret the exact path the model took to reach a conclusion, potentially masking underlying flaws in model behavior and biases in the training data. Furthermore, while the system demonstrates improved performance, it is not immune to hallucinations and other errors. If the model scales its reasoning on top of incorrect premises or intermediate steps, it may produce more convincing but ultimately flawed arguments. We hope that by providing a unified view of knowledge access along all dimensions (sequence, depth, experts) through the lens of attention, this work may help to understand and mitigate these downsides.

\subsection{Experiment Details}
\label{sec:experiment_details}

\begin{table}[H]
    \caption{Hyperparameters.}
    \label{tab:hparams}
    \begin{center}
    \begin{small}
    \begin{tabular}{llll}
        \toprule

        {Model} & \multicolumn{1}{c}{\textbf{Transformer}} & \multicolumn{1}{c}{\textbf{Dreamer $-$DA}} & \multicolumn{1}{c}{\textbf{Dreamer}} \\
        \cmidrule(lr){2-2}\cmidrule(lr){3-3}\cmidrule(lr){4-4}
        {Depth} & 8, 16, 24, 32 & 8, 16, 24, 32 & 8, 16, 24, 32 \\
        \midrule

        layers & = depth & 1 & 1 \\
        DR iterations & 1 & = depth & = depth \\
        EA expert intermediate size & 512 & 504 & 480, 480, 472, 472 \\
        EA experts & 32 & 257, 517, 778, 1039 & 266, 537, 822, 1097 \\

        \midrule

        EA active experts & \multicolumn{3}{l}{8} \\
        EA bias update rate & \multicolumn{3}{l}{1e-3} \\
        EA query/key size & \multicolumn{3}{l}{128} \\
        EA expert activation function & \multicolumn{3}{l}{$\operatorname{SwiGLU}$ \cite{shazeer2020glu}} \\

        SA query heads & \multicolumn{3}{l}{16} \\
        SA key/value heads & \multicolumn{3}{l}{8} \\
        SA query/key/value size & \multicolumn{3}{l}{128} \\
        SA context length & \multicolumn{3}{l}{4096} \\
        SA RoPE base & \multicolumn{3}{l}{10000} \\

        DA query heads & \multicolumn{3}{l}{1} \\
        DA key/value heads & \multicolumn{3}{l}{1} \\
        DA query/key/value size & \multicolumn{3}{l}{128} \\
        DA context length & \multicolumn{3}{l}{= depth} \\
        DA RoPE base & \multicolumn{3}{l}{500} \\

        SA/DA MoE experts & \multicolumn{3}{l}{= depth} \\
        SA/DA MoE active experts & \multicolumn{3}{l}{1} \\
        SA/DA MoE bias update rate & \multicolumn{3}{l}{1e-2} \\

        hidden size $h$ & \multicolumn{3}{l}{1024} \\

        default weight initialization & \multicolumn{3}{l}{$\sim\!\mathcal{N}(0, \sqrt{1 / (5 \cdot h)})$ \cite{nguyenTransformersTearsImproving2019b}} \\

        out-proj weight initialization & \multicolumn{3}{l}{$\sim\!\mathcal{N}(0, \sqrt{1 / (2.5 \cdot h \cdot \text{depth} \cdot \text{attn\_dims})})$, adapted from \cite{nguyenTransformersTearsImproving2019b},} \\
        & \multicolumn{3}{l}{\quad with attn\_dims = $\{3 \text{ if DA}, 2 \text{ else}\}$} \\

        tokenizer & \multicolumn{3}{l}{\href{https://huggingface.co/meta-llama/Llama-3.3-70B-Instruct}{Llama 3.3 70B Instruct}} \\

        optimizer & \multicolumn{3}{l}{AdamW \cite{kingmaAdamMethodStochastic2015,loshchilovDecoupledWeightDecay2019a} with $\beta$ = (0.9, 0.999), $\epsilon$ = 1e-8} \\
        max.\ learning rate (LR) & \multicolumn{3}{l}{4e-4} \\
        LR schedule & \multicolumn{3}{l}{constant after linear warmup for 500 steps} \\
        weight decay & \multicolumn{3}{l}{1e-3} \\
        max.\ gradient norm (clipped) & \multicolumn{3}{l}{0.5} \\
        batch size & \multicolumn{3}{l}{$2^{20}$ tokens} \\
        dtype & \multicolumn{3}{l}{bfloat16} \\

        accelerator & \multicolumn{3}{l}{NVIDIA H100 80GB HBM3} \\
        training runtimes (H100-hours) & \multicolumn{3}{l}{$\sim$1.4k, 1.7k, 2.1k, 2.6k (depth 8, 16, 24, 32; DA: +13\%), $\sim$34.5k total} \\

        \bottomrule
    \end{tabular}
    \end{small}
    \end{center}
\end{table}

\newpage

\begin{table}[H]
    \caption{Natural language reasoning \& math training datasets.}
    \label{tab:reasoning-math-datasets}
    \begin{center}
    \begin{small}
    \begin{tabular}{p{0.45\linewidth}p{0.1\linewidth}p{0.25\linewidth}}
        \toprule
        {\sc\textbf{Dataset}} & {\sc\textbf{Paper}} & {\sc\textbf{License}} \\
        \midrule
        \href{https://huggingface.co/datasets/EleutherAI/hendrycks_math}{EleutherAI/hendrycks\_math} & \cite{hendrycksMeasuringMathematicalProblem2021b} & \href{https://choosealicense.com/licenses/mit/}{MIT} \\
        \href{https://huggingface.co/datasets/bethgelab/CuratedThoughts}{bethgelab/CuratedThoughts} & \cite{hochlehnertCuratedThoughtsDataCuration2025} & - \\
            \quad (OpenR1-Math-220k-default, \\
            \quad OpenThoughts-114k-math-default) \\
        \href{https://huggingface.co/datasets/microsoft/orca-math-word-problems-200k}{microsoft/orca-math-word-problems-200k} & \cite{mitra2024orca} & \href{https://choosealicense.com/licenses/mit/}{MIT} \\
        \href{https://huggingface.co/datasets/TIGER-Lab/MathInstruct}{TIGER-Lab/MathInstruct} & \cite{yue2024mammoth} & \href{https://choosealicense.com/licenses/mit/}{MIT} \\
        \href{https://huggingface.co/datasets/openai/gsm8k}{openai/gsm8k} & \cite{cobbeTrainingVerifiersSolve2021b} & \href{https://choosealicense.com/licenses/mit/}{MIT} \\
        \href{https://huggingface.co/datasets/XinyaoHu/AMPS_khan}{XinyaoHu/AMPS\_khan} & - & \href{https://choosealicense.com/licenses/mit/}{MIT} \\
        \href{https://huggingface.co/datasets/open-thoughts/OpenThoughts2-1M}{open-thoughts/OpenThoughts2-1M} & \cite{guha2025openthoughts} & \href{https://choosealicense.com/licenses/apache-2.0/}{Apache 2.0} \\
        \href{https://huggingface.co/datasets/AI-MO/NuminaMath-CoT}{AI-MO/NuminaMath-CoT} & \cite{liNuminaMathLargestPublic2024} & \href{https://choosealicense.com/licenses/apache-2.0/}{Apache 2.0} \\
        \href{https://huggingface.co/datasets/facebook/natural_reasoning}{facebook/natural\_reasoning} & \cite{yuan2025naturalreasoning} & \href{https://spdx.org/licenses/CC-BY-NC-4.0}{CC-BY-NC-4.0} \\
        \href{https://huggingface.co/datasets/nvidia/OpenMathInstruct-2}{nvidia/OpenMathInstruct-2} & \cite{toshniwalOpenMathInstruct2AcceleratingAI2025} & \href{https://choosealicense.com/licenses/cc-by-4.0/}{CC-BY-4.0} \\
        \href{https://huggingface.co/datasets/nvidia/OpenMathReasoning}{nvidia/OpenMathReasoning} (cot) & \cite{moshkov2025aimo} & \href{https://choosealicense.com/licenses/cc-by-4.0/}{CC-BY-4.0} \\
        \href{https://huggingface.co/datasets/zwhe99/DeepMath-103K}{zwhe99/DeepMath-103K} & \cite{he2025deepmath} & \href{https://choosealicense.com/licenses/mit/}{MIT} \\
        \href{https://huggingface.co/datasets/ibivibiv/math_instruct}{ibivibiv/math\_instruct} & - & - \\
        \href{https://huggingface.co/datasets/nvidia/Nemotron-Pretraining-SFT-v1}{nvidia/Nemotron-Pretraining-SFT-v1} & \cite{basant2025nvidia} & \multirow{2}{*}{\href{https://huggingface.co/datasets/nvidia/Nemotron-Pretraining-SFT-v1/blob/main/LICENSE.md}{\begin{tabular}{@{}l@{}}
            NVIDIA Data Agreement \\
            \quad for Model Training
        \end{tabular}}} \\
        \quad (Nemotron-SFT-MATH) \\
        \href{https://huggingface.co/datasets/ajibawa-2023/Maths-College}{ajibawa-2023/Maths-College} & - & \href{https://choosealicense.com/licenses/apache-2.0/}{Apache 2.0} \\
        \bottomrule
    \end{tabular}
    \end{small}
    \end{center}
\end{table}

\begin{table}[H]
    \caption{Natural language reasoning \& math test benchmarks.}
    \label{tab:reasoning-math-benchmarks}
    \begin{center}
    \begin{small}
    \begin{tabular}{p{0.45\linewidth}p{0.1\linewidth}p{0.25\linewidth}}
        \toprule
        {\sc\textbf{Benchmark}} & {\sc\textbf{Paper}} & {\sc\textbf{License}} \\
        \midrule
        \href{https://huggingface.co/datasets/allenai/math_qa}{MathQA} & \cite{amini2019mathqa} & \href{https://choosealicense.com/licenses/apache-2.0/}{Apache 2.0} \\
        \href{https://huggingface.co/datasets/openai/gsm8k}{GSM8K} & \cite{cobbeTrainingVerifiersSolve2021b} & \href{https://choosealicense.com/licenses/mit/}{MIT} \\
        \href{https://huggingface.co/datasets/EleutherAI/hendrycks_math}{MATH} & \cite{hendrycksMeasuringMathematicalProblem2021b} & \href{https://choosealicense.com/licenses/mit/}{MIT} \\
        \href{https://huggingface.co/datasets/cais/mmlu}{MMLU} (math: & \cite{hendrycks2021measuring} & \href{https://choosealicense.com/licenses/mit/}{MIT} \\
            \quad abstract\_algebra, \\
            \quad elementary\_mathematics, \\
            \quad high\_school\_mathematics, \\
            \quad college\_mathematics, \\
            \quad high\_school\_statistics) \\
        \href{https://huggingface.co/datasets/ajibawa-2023/Maths-College}{Maths-College} & - & \href{https://choosealicense.com/licenses/apache-2.0/}{Apache 2.0} \\
        \bottomrule
    \end{tabular}
    \end{small}
    \end{center}
\end{table}

\begin{table}[H]
    \caption{Knowledge \& natural language training datasets.}
    \label{tab:knowledge-language-datasets}
    \begin{center}
    \begin{small}
    \begin{tabular}{p{0.45\linewidth}p{0.1\linewidth}p{0.25\linewidth}}
        \toprule
        {\sc\textbf{Dataset}} & {\sc\textbf{Paper}} & {\sc\textbf{License}} \\
        \midrule
        \href{https://huggingface.co/datasets/nvidia/Nemotron-CC-v2}{nvidia/Nemotron-CC-v2} & \cite{nvidiaNVIDIANemotronNano2025a} & \multirow{2}{*}{\href{https://huggingface.co/datasets/nvidia/Nemotron-CC-v2/blob/main/LICENSE.md}{\begin{tabular}{@{}l@{}}
            NVIDIA Data Agreement \\
            \quad for Model Training
        \end{tabular}}} \\
        \quad (High-Quality-Synthetic) \\
        \bottomrule
    \end{tabular}
    \end{small}
    \end{center}
\end{table}

\begin{table}[H]
    \caption{Knowledge \& natural language test benchmarks.}
    \label{tab:knowledge-language-benchmarks}
    \begin{center}
    \begin{small}
    \begin{tabular}{p{0.45\linewidth}p{0.1\linewidth}p{0.25\linewidth}}
        \toprule
        {\sc\textbf{Benchmark}} & {\sc\textbf{Paper}} & {\sc\textbf{License}} \\
        \midrule
        \href{https://huggingface.co/datasets/Rowan/hellaswag}{HellaSwag} & \cite{zellers2019hellaswag} & \href{https://choosealicense.com/licenses/mit/}{MIT} \\
        \href{https://huggingface.co/datasets/cais/mmlu}{MMLU} & \cite{hendrycks2021measuring} & \href{https://choosealicense.com/licenses/mit/}{MIT} \\
        \href{https://huggingface.co/datasets/baber/piqa}{PIQA} & \cite{Bisk2020} & - \\
        \href{https://huggingface.co/datasets/mandarjoshi/trivia_qa}{TriviaQA} & \cite{JoshiTriviaQA2017} & - \\
        \bottomrule
    \end{tabular}
    \end{small}
    \end{center}
\end{table}

\subsection{Long Context Efficiency}
\label{sec:long-context-efficiency}

\begin{table}[H]
    \footnotesize
    \caption{Inference FLOPs per token (in billions) for Dreamer and Transformer (depth 16) at varying context lengths, decomposed into Dreamer's components (SA, DA, EA).}
    \label{tab:efficiency-flops}
    \begin{center}
    \begin{small}
    \begin{tabular}{rccccc}
        \toprule
        \textsc{\textbf{Context Length}} & \textsc{\textbf{Transformer}} & \textsc{\textbf{Dreamer}} & \textsc{\textbf{SA}} & \textsc{\textbf{DA}} & \textsc{\textbf{EA}} \\
        \midrule
        1{,}024 & 0.939 & 0.941 & 0.273 & 0.021 & 0.384 \\
        16{,}384 & 1.951 & 1.954 & 1.286 & 0.021 & 0.384 \\
        131{,}072 & 9.512 & 9.514 & 8.846 & 0.021 & 0.384 \\
        1{,}048{,}576 & 69.993 & 69.996 & 69.328 & 0.021 & 0.384 \\
        \bottomrule
    \end{tabular}
    \end{small}
    \end{center}
\end{table}

\begin{table}[H]
    \footnotesize
    \caption{Inference peak memory (in GB) for Dreamer and Transformer (depth 16) at varying context lengths, decomposed into Dreamer's components (SA, DA, EA). Peak memory includes parameters, activations, and kv-cache.}
    \label{tab:efficiency-memory}
    \begin{center}
    \begin{small}
    \begin{tabular}{rccccc}
        \toprule
        \textsc{\textbf{Context Length}} & \textsc{\textbf{Transformer}} & \textsc{\textbf{Dreamer}} & \textsc{\textbf{SA}} & \textsc{\textbf{DA}} & \textsc{\textbf{EA}} \\
        \midrule
        1{,}024 & 2.452 & 2.461 & 0.281 & 0.018 & 1.593 \\
        16{,}384 & 3.585 & 3.585 & 1.405 & 0.018 & 1.584 \\
        131{,}072 & 12.040 & 12.041 & 9.861 & 0.018 & 1.584 \\
        1{,}048{,}576 & 79.688 & 79.689 & 77.507 & 0.018 & 1.584 \\
        \bottomrule
    \end{tabular}
    \end{small}
    \end{center}
\end{table}

\subsection{Benchmark Results}

\begin{figure}[H]
    \centering
    \begin{subfigure}[b]{\linewidth}
        \centerline{\includegraphics[width=\linewidth]{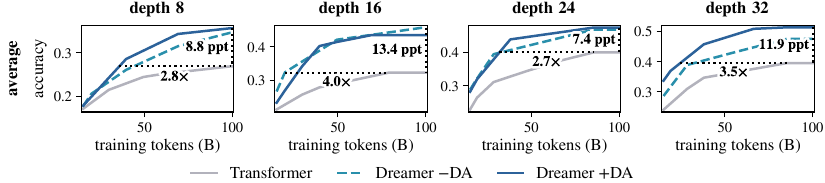}}%
    \end{subfigure}%
    \caption{Intermediate reasoning benchmark results (0-shot) during training, with FLOP-/parameter-/memory-matched architectures per depth. Dreamer consistently outperforms modern Transformers.}
    \label{fig:reasoning-math-results}
\end{figure}

\subsection{Expert Attention}

\begin{figure}[H]
    \centering
    \centerline{\includegraphics[width=\linewidth]{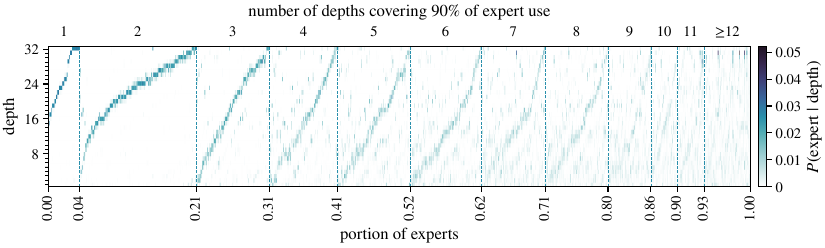}}%
    \caption{
        Distribution of experts per depth. Experts are sorted like in \cref{fig:ea-map-depth-prob}.
        Observations: Higher depths tend to use more depth-specialized experts.
        However, there are some exceptions, like the second to last depth, which also use some specific depth-generalized experts.
    }
    \label{fig:ea-map-expert-prob}
\end{figure}

In addition to \cref{fig:ea-map-depth-prob}, which shows $P(\text{depth}|\text{expert})$ for EA, \cref{fig:ea-map-expert-prob} shows $P(\text{expert}|\text{depth})$.

\end{document}